\documentclass[lettersize,journal]{IEEEtran}

\usepackage{cite}
\usepackage{amsmath,amssymb,amsfonts}
\usepackage{graphicx}
\usepackage{dblfloatfix}
\usepackage{textcomp}
\usepackage{xcolor}
\usepackage{booktabs}
\usepackage{multirow}
\usepackage{array}
\usepackage{url}
\usepackage{balance}
\usepackage{tikz}
\usepackage{placeins}
\DeclareMathOperator*{\argmin}{arg\,min}
\DeclareMathOperator{\clip}{clip}

\newcommand{\safeincludegraphics}[2][]{%
  \IfFileExists{#2}{\includegraphics[#1]{#2}}{%
    \fbox{\parbox[c][0.72in][c]{0.90\linewidth}{\centering Missing image:\\\detokenize{#2}}}%
  }%
}

\begin{document}

\title{GenVid2Robot: From Video Generation to Robot Manipulation via Rigid-Geometric Consistency}

\author{Haohui Huang,~\IEEEmembership{Member,~IEEE},
Xi Yuan,
Panpan Liao,
Tao Teng,~\IEEEmembership{Member,~IEEE},
Chenguang Yang,~\IEEEmembership{Fellow,~IEEE},
Jing Guo,~\IEEEmembership{Member,~IEEE},
and Yi Guo,~\IEEEmembership{Member,~IEEE}%
\thanks{Corresponding author: Tao Teng.}%
\thanks{Haohui Huang, Xi Yuan, Panpan Liao, and Jing Guo are with the School of Automation, Guangdong University of Technology, Guangzhou, China.}%
\thanks{Tao Teng is with the University of Liverpool, Liverpool, U.K. (e-mail: Tao.Teng@liverpool.ac.uk).}%
\thanks{Chenguang Yang is with the Department of Computing, The Hong Kong Polytechnic University, Hong Kong, China.}%
\thanks{Yi Guo is with the State Key Laboratory of Submarine Geoscience, School of Automation and Intelligent Sensing, Shanghai Jiao Tong University, Shanghai 200240, China.}%
}

\markboth{IEEE Transactions Journal Manuscript}%
{Huang \MakeLowercase{\textit{et al.}}: GenVid2Robot: From Video Generation to Robot Manipulation via Rigid-Geometric Consistency}

\maketitle

\begin{abstract}
Generated videos provide useful visual motion priors for robot manipulation, but their visual plausibility does not imply physical executability. A generated video usually lacks metric geometry, grasp grounding, robot kinematic feasibility, and execution-time feedback, which makes direct trajectory replay unreliable in real-world manipulation. This paper presents GenVid2Robot, a rigid-geometric consistency framework that converts generated video motion into executable real-robot manipulation trajectories. Given an initial RGB-D observation and a task instruction, GenVid2Robot samples task-relevant semantic anchors from the real first frame, tracks these anchors through generated video candidates, and verifies whether the resulting 2D motion can be explained by first-frame RGB-D anchors under a sparse relative $SE(3)$ model. In this way, generated videos are treated as uncertain visual motion hypotheses rather than direct robot demonstrations. Only geometrically consistent motion is transferred to the robot. The accepted relative motion is then applied to the real grasp-time TCP pose selected by mask-constrained grasping, producing a grasp-conditioned execution trajectory that is consistent with both the visual motion prior and the physical grasp configuration. To reduce execution mismatch caused by RGB-D noise, calibration residuals, and small contact-induced displacement, a bounded depth-compensation module corrects local depth-direction errors without assuming full online replanning. Real-robot experiments demonstrate that GenVid2Robot improves the reliability of generated-video-guided manipulation by grounding visual motion priors with sparse metric geometry, grasp constraints, robot feasibility checking, and bounded execution feedback.
\end{abstract}


\section{Introduction}\label{sec:introduction}

Robots are increasingly expected to perform manipulation tasks from visual observations and high-level language instructions, rather than relying on manually designed trajectories for every object, scene, and task variation. Tasks such as pouring water, lifting a lid, delivering a tool, or sweeping an object require the robot to infer task-relevant object motion, establish a feasible grasp, and execute a physically valid trajectory under sensing and calibration uncertainty. Although reinforcement learning, imitation learning, and diffusion-based visuomotor policies have achieved strong results in robot manipulation, their performance is still closely tied to the availability, diversity, and embodiment compatibility of physical demonstrations or robot interaction data \cite{gu2017deep,zhao2023learning,chi2023diffusion}.

Foundation models provide a promising way to reduce task-specific supervision by introducing language, semantic, and spatial priors into robot manipulation. Large multimodal models can generate plans, reason about affordances, produce executable code, or define spatial constraints for manipulation \cite{ahn2022can,driess2023palm,zitkovich2023rt2,liang2023code,huang2023voxposer,huang2024rekep}. However, these high-level priors still need to be converted into continuous robot motion that is geometrically grounded, grasp-compatible, and physically executable. This conversion remains difficult when no object CAD model, object-specific demonstration, or pre-defined task trajectory is available.

Generated videos provide another source of manipulation priors. Given an initial image and a task instruction, an image-to-video model can synthesize a temporally ordered visual sequence that illustrates how an object may move. Recent generated-video-based manipulation systems explore this idea by extracting object trajectories or actionable visual motion from generated videos and transferring them to robot execution \cite{patel2025robotic,li2025novaflow}. These methods suggest that generated videos can provide useful object-centric motion cues without collecting physical demonstrations.

However, a generated video is not a robot trajectory. It mainly provides image-space visual dynamics, while real robot execution requires metric 3D motion, rigid-body consistency, grasp feasibility, workspace compatibility, and execution-time correction. A visually plausible generated video may still contain drifting correspondences, non-rigid artifacts, inconsistent depth evolution, or motions that cannot be realized by the robot. Directly retargeting such visual motion to a robot may therefore lead to unstable motion recovery, incorrect grasp-relative execution, or physical task failure.

The central observation of this paper is that the main gap between video generation and robot manipulation is not only semantic, but also geometric and physical. Generated videos should be treated as uncertain two-dimensional motion hypotheses rather than directly executable demonstrations. Instead of directly imitating generated pixels or relying on an end-to-end mapping from generated videos to robot actions, we introduce rigid-geometric consistency as an intermediate validation mechanism. The key requirement is that the tracked two-dimensional motion in a generated video must be explainable by a common sparse rigid $SE(3)$ transformation of three-dimensional anchors reconstructed from the real first RGB-D observation. Only when this consistency condition is satisfied is the recovered motion allowed to enter the robot execution pipeline.

To this end, we propose \textbf{GenVid2Robot}, a framework for converting video-generation-based visual motion priors into real-robot manipulation trajectories. GenVid2Robot first initializes task-relevant semantic anchors from the initial RGB-D observation using part prompts and object masks, and back-projects these anchors into metric 3D space. It then tracks the corresponding two-dimensional anchors in generated video candidates and estimates frame-wise relative anchor-set motion using PnP/RANSAC. A reprojection-based rigid-geometric consistency test rejects generated motions with unstable correspondences, visual drift, or non-rigid behavior. The accepted relative motion is transformed into the robot base frame and applied to the real grasp-time TCP pose, yielding a grasp-conditioned TCP trajectory rather than a direct copy of an object-center trajectory. Mask-constrained grasp selection, IK feasibility checking, trajectory smoothing, and bounded RealSense depth correction are further used to improve physical feasibility and execution robustness.

The contribution of GenVid2Robot is not a new vision-language model, point tracker, PnP solver, or grasp detector. Rather, the contribution lies in organizing these components into a physically interpretable video-to-robot manipulation pipeline in which generated visual motion is first constrained by sparse rigid geometry before it is allowed to affect robot execution. This design preserves the flexibility of generated video priors while preventing geometrically inconsistent visual motion from being transferred directly to the robot.

The main contributions are summarized as follows:
\begin{itemize}
    \item \textbf{Video-generation-to-robot manipulation framework}: We propose GenVid2Robot, a pipeline that converts video-generation-based visual motion priors into executable robot manipulation trajectories without requiring object CAD models, offline object scanning, or pre-collected object-specific physical demonstrations.

    \item \textbf{Rigid-geometric consistency for generated visual motion}: We introduce a consistency mechanism that tests whether tracked two-dimensional semantic anchors in a generated video can be explained by a common sparse rigid $SE(3)$ motion of first-frame RGB-D anchors. This mechanism bridges generated image-space dynamics and physically interpretable 3D relative motion.

    \item \textbf{Grasp-conditioned TCP trajectory induction}: We convert the accepted relative motion into a robot-executable trajectory by applying it to the real grasp-time TCP pose, rather than directly copying object-center coordinates. Mask-constrained grasping, IK feasibility checking, and bounded RealSense depth correction further improve grasp feasibility and execution robustness.

    \item \textbf{Real-robot evaluation and diagnostic analysis}: We evaluate GenVid2Robot on an RM75 manipulator across pouring, lifting, tool delivery, and sweeping tasks, and compare it with representative constraint-based, generated-video-trajectory, and actionable-flow baselines. Ablation and diagnostic studies analyze rigid-geometric filtering, semantic anchor selection, layout perturbation, and depth-corrected execution.
\end{itemize}

\section{Related Work}\label{sec:related_work}

\subsection{Robot Priors for Manipulation}

Robot manipulation has been widely studied through reinforcement learning, imitation learning, and visuomotor policy learning. Reinforcement-learning methods can acquire manipulation skills through interaction, but they often require extensive robot-environment data and carefully designed rewards \cite{gu2017deep}. Imitation learning and diffusion-based policies reduce exploration by learning from demonstrations, as shown by ACT-style fine-grained manipulation policies and Diffusion Policy \cite{zhao2023learning,chi2023diffusion}. However, their generalization to new objects, task variants, and robot embodiments is still constrained by the coverage of physical demonstrations.

Foundation models provide another way to introduce task priors. SayCan grounds language planning with affordance values \cite{ahn2022can}; PaLM-E and RT-2 connect multimodal representations to embodied reasoning or robot actions \cite{driess2023palm,zitkovich2023rt2}; and Code as Policies generates executable robot programs from language \cite{liang2023code}. VoxPoser and ReKep further use large models to produce spatial value maps or relational keypoint constraints for manipulation \cite{huang2023voxposer,huang2024rekep}. These methods show the value of semantic and spatial priors, but the resulting plans or constraints still need to be converted into continuous, geometrically grounded, grasp-compatible, and physically executable robot motion. GenVid2Robot focuses on this conversion problem by treating generated videos as visual motion priors and validating their motion through sparse rigid geometry before robot execution.

\subsection{Generated Videos and Sparse Motion Grounding}

Generated videos provide an emerging source of object-centric motion priors. RIGVid generates candidate videos, filters them with a VLM, extracts 6D object trajectories, and retargets the motion to a robot without physical demonstrations \cite{patel2025robotic}. NovaFlow further converts generated video motion into actionable object flow and robot actions \cite{li2025novaflow}. These works demonstrate the potential of generated videos for robot manipulation. However, a visually plausible generated video may still contain drifting correspondences, non-rigid artifacts, inaccurate depth evolution, or motions that violate robot workspace and grasp constraints. Therefore, generated videos should be treated as uncertain visual motion hypotheses rather than directly executable robot trajectories.

To obtain metric motion from video, visual correspondences must be connected to 3D geometry. Dense optical flow and point trackers provide image-space motion cues, while CoTracker improves long-term point association by jointly tracking multiple points \cite{karaev2024cotracker}. Classical PnP estimates camera-object motion from 2D--3D correspondences and can be combined with RANSAC to reject outliers \cite{fischler1981random,lepetit2009epnp}. Modern 6D pose trackers such as FoundationPose provide strong object pose estimation, but they usually require CAD models or reference views at test time \cite{wen2024foundationpose}. In contrast, GenVid2Robot uses first-frame RGB-D semantic anchors as sparse 3D references and estimates relative anchor-set motion from generated video tracks. The recovered motion is interpreted as a relative sparse motion prior rather than a CAD-defined canonical object pose.

\subsection{Grasping and Execution Feedback}

Video-guided manipulation also requires functional grasping and execution-time correction. Data-driven grasp synthesis methods estimate grasp poses from depth or point clouds, and AnyGrasp provides dense full-DoF grasp candidates for unseen objects under depth noise \cite{bohg2014data,fang2023anygrasp}. However, a high-confidence grasp may still be task-incompatible if it is not aligned with the manipulated object region or the recovered motion.

GenVid2Robot uses mask-constrained grasping to restrict grasp candidates to task-relevant object regions and induces the robot TCP trajectory from the selected grasp-time TCP pose and the verified relative motion. This grasp-conditioned formulation avoids directly copying object-center coordinates or generated image-space trajectories. During execution, RealSense depth feedback provides bounded local compensation for depth mismatch caused by generated-video geometry, RGB-D noise, calibration residuals, and contact-induced displacement. This feedback is intentionally limited to local depth correction and is not treated as full online visual servoing or task-level replanning.

\section{Problem Formulation}\label{sec:problem_formulation}

Given an initial RGB-D observation and a natural-language task instruction, the objective of GenVid2Robot is to convert video-generation-based visual motion priors into executable robot manipulation trajectories without requiring object CAD models, offline object scanning, or pre-collected object-specific physical demonstrations. The initial observation is
\begin{equation}
    \mathcal{O}_0=
    \left\{
    \mathbf{I}_0,\mathbf{D}_0,\mathbf{K},\mathbf{T}_{b,c},l
    \right\},
    \label{eq:initial_observation}
\end{equation}
where $\mathbf{I}_0$ and $\mathbf{D}_0$ are the aligned RGB image and metric depth map, $\mathbf{K}$ is the camera intrinsic matrix, $\mathbf{T}_{b,c}\in SE(3)$ is the calibrated camera-to-base transform, and $l$ is the task instruction.

A video generation model produces candidate visual motion hypotheses from the initial image and instruction:
\begin{equation}
    \hat{\mathcal{V}}_n
    =
    \mathcal{M}_{vid}(\mathbf{I}_0,l),
    \quad n=1,\ldots,N_v .
    \label{eq:generated_video_candidates}
\end{equation}
Each $\hat{\mathcal{V}}_n$ provides image-space temporal motion, but it does not directly provide metric geometry, robot kinematics, grasp feasibility, or execution feedback. Therefore, a generated video is treated as an uncertain 2D motion hypothesis rather than as an executable robot trajectory.

GenVid2Robot initializes $K$ task-relevant semantic anchors in the first RGB-D frame and back-projects them into metric 3D space:
\begin{equation}
\begin{aligned}
    \mathbf{q}_i^{(0)}
    &=
    \begin{bmatrix}
    u_i^{(0)} & v_i^{(0)}
    \end{bmatrix}^{\top},\\
    \mathbf{p}_{i}^{c,0}
    &=
    z_i^{(0)}
    \mathbf{K}^{-1}
    \begin{bmatrix}
    u_i^{(0)}\\
    v_i^{(0)}\\
    1
    \end{bmatrix},
    \quad
    z_i^{(0)}
    =
    \mathbf{D}_0
    \left(u_i^{(0)},v_i^{(0)}\right),\\
    \mathcal{P}_{c}^{0}
    &=
    \left\{
    \bar{\mathbf{p}}_{i}^{c,0}
    \right\}_{i=1}^{K},
    \quad
    \bar{\mathbf{p}}_{i}^{c,0}
    =
    \begin{bmatrix}
    (\mathbf{p}_{i}^{c,0})^{\top} & 1
    \end{bmatrix}^{\top}.
\end{aligned}
    \label{eq:first_frame_backprojection}
\end{equation}
The anchor set $\mathcal{P}_{c}^{0}$ is defined directly in the first camera frame. No CAD-defined canonical object frame or explicit object coordinate system is assumed. Thus, the recovered motion is interpreted as the relative rigid motion of the first-frame sparse anchor set, rather than as a canonical object 6D pose trajectory.

For a generated video candidate $\hat{\mathcal{V}}_n$, let $\mathbf{q}_{i,n}^{(t)}$ be the tracked 2D location of anchor $i$ at frame $t$. The rigid-geometric consistency requirement is that the tracked 2D motion should be explainable by a common frame-wise relative transformation $\Delta\mathbf{T}_{c,n}^{(t)}\in SE(3)$ applied to the first-frame 3D anchors:
\begin{equation}
    \mathbf{q}_{i,n}^{(t)}
    \approx
    \Pi
    \left(
    \mathbf{K},
    \Delta\mathbf{T}_{c,n}^{(t)}
    \bar{\mathbf{p}}_{i}^{c,0}
    \right),
    \quad
    i\in\mathcal{I}_{n,t},
    \label{eq:rigid_geometric_consistency_condition}
\end{equation}
where $\Pi(\cdot)$ is the camera projection function and $\mathcal{I}_{n,t}$ is the inlier set. A generated motion is considered geometrically transferable only when its tracked image-space motion can be explained by sparse rigid $SE(3)$ motion with sufficiently low reprojection residuals.

The recovered relative motion is transformed from the camera frame to the robot base frame by
\begin{equation}
    \Delta\mathbf{T}_{b,n}^{(t)}
    =
    \mathbf{T}_{b,c}
    \Delta\mathbf{T}_{c,n}^{(t)}
    \mathbf{T}_{c,b},
    \quad
    \mathbf{T}_{c,b}
    =
    \mathbf{T}_{b,c}^{-1}.
    \label{eq:relative_motion_base}
\end{equation}
After rigid-geometric validation, the accepted relative motion is denoted as $\Delta\bar{\mathbf{T}}_{b}^{(t)}$. Given the real grasp-time TCP pose $\mathbf{T}_{b,tcp}^{(0)}$, the nominal robot TCP trajectory is induced by
\begin{equation}
    \mathbf{T}_{b,tcp}^{nom,(t)}
    =
    \Delta\bar{\mathbf{T}}_{b}^{(t)}
    \mathbf{T}_{b,tcp}^{(0)},
    \quad
    t=1,\ldots,T .
    \label{eq:tcp_trajectory_problem}
\end{equation}
This formulation emphasizes that the robot does not directly copy generated image-space trajectories or object-center coordinates. Instead, the generated visual motion is first converted into validated relative rigid motion, and the executable TCP trajectory is then induced from the selected grasp-time TCP pose.

The final command trajectory must satisfy visual-motion consistency, grasp compatibility, robot feasibility, and bounded execution correction:
\begin{equation}
    \mathcal{T}_{cmd}
    =
    \left\{
    \mathbf{T}_{cmd}^{(t)}
    \right\}_{t=1}^{T}
    =
    \mathcal{F}
    \left(
    \left\{
    \Delta\bar{\mathbf{T}}_{b}^{(t)}
    \right\}_{t=1}^{T},
    \mathbf{T}_{b,tcp}^{(0)},
    \mathcal{C}_{robot},
    \mathcal{D}_{rs}
    \right),
    \label{eq:final_command_trajectory}
\end{equation}
where $\mathcal{C}_{robot}$ denotes robot kinematic and feasibility constraints, and $\mathcal{D}_{rs}$ denotes the execution-time RealSense depth stream.

\subsection{Geometric and Physical Assumptions}

GenVid2Robot relies on three assumptions. First, over a short manipulation horizon, the task-relevant object region can be approximated by a sparse rigid anchor set, so a transferable generated video should produce 2D tracks that are explainable by a common relative $SE(3)$ motion. Second, after grasping, the robot TCP and the manipulated object approximately form a local rigid relation, so the TCP trajectory should be induced from the grasp-time TCP pose and the recovered relative motion rather than by directly following object-center coordinates. Third, generated-video geometry and real-world execution may differ because of visual artifacts, RGB-D noise, calibration residuals, and contact-induced displacement; bounded depth feedback is therefore used only as local execution compensation.

These assumptions imply three consistency principles: projection consistency between generated 2D tracks and first-frame 3D anchors, rigid-body consistency of the accepted relative $SE(3)$ motion, and grasp-relative consistency between the selected TCP pose and the induced robot trajectory. Thus, GenVid2Robot does not imitate generated pixels directly. It first converts generated visual dynamics into physically interpretable relative motion and then transfers the verified motion to robot execution.

\section{Methodology}\label{sec:methodology}

\begin{figure*}[!t]
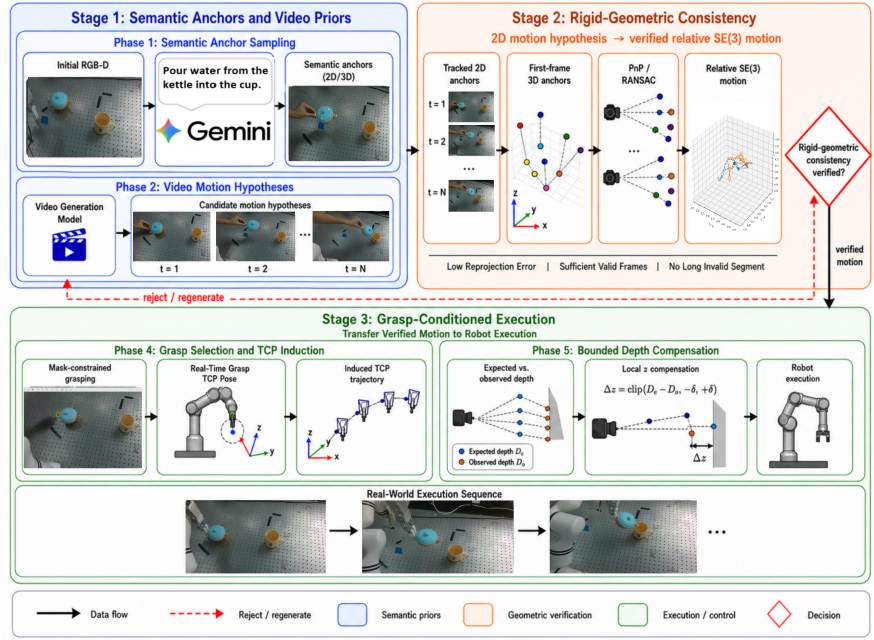

    \centering
    \safeincludegraphics[width=0.9\textwidth]{12.jpg}
    \caption{Technical route of GenVid2Robot. The framework consists of three stages and five phases. Stage~1 samples semantic anchors from the first RGB-D observation and generates candidate video motion hypotheses. Stage~2 performs rigid-geometric consistency verification by combining tracked 2D anchors, first-frame 3D anchors, PnP/RANSAC, and reprojection checks. Stage~3 selects a mask-constrained grasp, induces a grasp-conditioned TCP trajectory, and applies bounded depth compensation before real-robot execution. Candidates that fail the rigid-geometric consistency check are rejected or regenerated.}
    \label{fig:pipeline}
\end{figure*}

Fig.~\ref{fig:pipeline} summarizes the GenVid2Robot pipeline. The method contains three stages and five phases. Stage~1 constructs semantic anchors and generated video motion hypotheses from the initial RGB-D observation and task instruction. Stage~2 verifies whether the tracked 2D anchor motion in a generated video can be explained by a sparse rigid $SE(3)$ motion of first-frame 3D anchors. Stage~3 transfers the verified relative motion to the robot through grasp-conditioned TCP induction and bounded depth compensation. Thus, generated videos are used as visual motion hypotheses, and only geometrically consistent motion is allowed to enter robot execution.

\subsection{Stage 1: Semantic Anchors and Video Priors}\label{sec:stage1}

\subsubsection{Phase 1: Semantic Anchor Sampling}\label{sec:semantic_anchor_sampling}

Given the initial RGB-D observation $\mathcal{O}_0$ and task instruction $l$, Phase~1 constructs task-relevant semantic anchors in the real first frame. A VLM proposes part-level prompts associated with the task, such as handle, body, spout, and lid in a pouring task. These prompts are grounded in the first RGB image and segmented by SAM \cite{kirillov2023segment}, producing part masks $\{M_m\}_{m=1}^{M}$. The purpose is to initialize anchors on semantically meaningful and geometrically stable object regions rather than arbitrary pixels or manually annotated keypoints.

For each part mask, pixels with invalid depth, boundary uncertainty, or overly dense neighboring samples are removed. We denote the valid sampling region as $\Omega_m$, where each pixel $(u,v)\in\Omega_m$ satisfies $(u,v)\in M_m$, $\mathbf{D}_0(u,v)>0$, and $d((u,v),\partial M_m)>\epsilon_b$. The depth-aware feature and part-wise K-Means centers are defined as
\begin{equation}
\begin{aligned}
    \phi(u,v)
    &=
    \left[
    u/W,\;
    v/H,\;
    \mathbf{D}_0(u,v)/d_s
    \right]^{\top},\\
    \{\mathbf{c}_{m,k}\}_{k=1}^{K_m}
    &=
    \operatorname{KMeans}
    \left(
    \{\phi(u,v)\mid (u,v)\in\Omega_m\},K_m
    \right).
\end{aligned}
    \label{eq:method_kmeans_anchor_sampling}
\end{equation}
Here, $\partial M_m$ is the mask boundary, $\epsilon_b$ is the boundary margin, $W$ and $H$ are the image size, and $d_s$ is a depth normalization constant. In our implementation, $K_m=5$ is used for each valid part mask.

The anchor corresponding to each cluster center is selected as the valid mask pixel closest to that center:
\begin{equation}
    \mathbf{q}_{m,k}^{(0)}
    =
    \arg\min_{(u,v)\in\Omega_m}
    \left\|
    \phi(u,v)-c_{m,k}
    \right\|_2 ,
    \quad
    k=1,\ldots,K_m .
    \label{eq:anchor_selection}
\end{equation}
Nearby duplicate anchors are suppressed to improve spatial coverage. The selected 2D anchors are back-projected into first-frame 3D anchors using Eq.~\eqref{eq:first_frame_backprojection}, forming the sparse metric reference $\mathcal{P}_{c}^{0}$ for subsequent rigid-geometric verification.

\begin{figure}[!t]
    \centering
    \includegraphics[width=0.92\columnwidth]{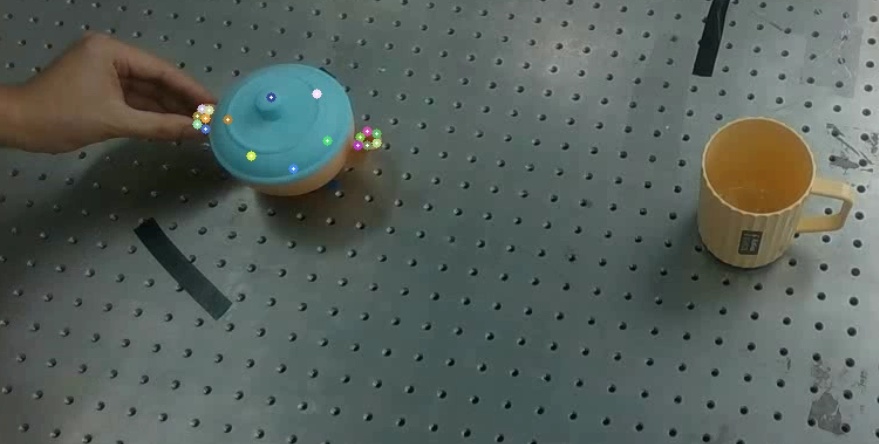}
    \vspace{-1mm}
    \caption{Part-prompt-guided semantic anchors. SAM-segmented part masks and depth-aware K-Means produce sparse task-relevant anchors for first-frame RGB-D reconstruction and rigid-geometric consistency verification.}
    \label{fig:keypoints}
    \vspace{-2mm}
\end{figure}

\subsubsection{Phase 2: Video Motion Hypotheses}\label{sec:video_motion_hypothesis}

The image-to-video model $\mathcal{M}_{vid}$ generates multiple candidate videos from the real first image and task instruction, as defined in Eq.~\eqref{eq:generated_video_candidates}. In our implementation, Kling 1.6 generates five candidates for each task. A VLM-based semantic screening step removes candidates with inconsistent task semantics, object identity, object shape, or overall motion direction.

The remaining candidates are treated as 2D visual motion hypotheses. For each candidate, CoTracker \cite{karaev2024cotracker} propagates the initialized semantic anchors through the generated frames:
\begin{equation}
    \left(
    \mathbf{q}_{i,n}^{(t)},\alpha_{i,n}^{(t)}
    \right)
    =
    \mathcal{C}_{track}
    \left(
    \hat{\mathcal{V}}_n,
    \mathbf{q}_{i}^{(0)}
    \right),
    \quad
    t=1,\ldots,T ,
    \label{eq:cotracker_tracking}
\end{equation}
where $\mathcal{C}_{track}$ is the point tracker and $\alpha_{i,n}^{(t)}$ is the visibility or confidence score of anchor $i$ in candidate $n$ at frame $t$. Only anchors initialized from the object region are used for motion recovery; generated hands or contextual elements are ignored.

\subsection{Stage 2: Rigid-Geometric Consistency}\label{sec:stage2}

\subsubsection{Phase 3: Rigid-Geometric Consistency Verification}\label{sec:phase3}

Phase~3 converts generated 2D anchor tracks into physically interpretable relative 3D motion. Given first-frame 3D anchors and their tracked 2D locations in a generated video, GenVid2Robot estimates a frame-wise sparse rigid motion and verifies whether the generated image-space motion can be explained by a common $SE(3)$ transformation under camera projection. Candidates that fail this verification are rejected or regenerated before robot execution.

\paragraph{Visibility and degeneracy handling.}
Before PnP/RANSAC, anchors with low tracking confidence, out-of-image locations, invalid object association, or unreliable visibility are removed. PnP is applied only when at least four valid anchors remain and their first-frame 3D points are not nearly collinear or poorly distributed. Let $\mathcal{T}_{n}^{valid}$ be the valid frame set of the $n$-th candidate. We compute
\begin{equation}
\begin{aligned}
    r_n^{valid}
    &=
    \frac{|\mathcal{T}_{n}^{valid}|}{T},\\
    L_n^{invalid}
    &=
    \max_{\tau\subseteq
    \{1,\ldots,T\}\setminus\mathcal{T}_{n}^{valid}}
    |\tau|.
\end{aligned}
    \label{eq:valid_ratio_invalid_segment}
\end{equation}
Frames with insufficient or degenerate correspondences are marked invalid and excluded from reprojection-error computation. If $r_n^{valid}$ is too low or $L_n^{invalid}$ is too large, the whole generated candidate is rejected.

\paragraph{PnP/RANSAC-based relative motion estimation.}
For valid frames, the first-frame RGB-D anchors provide metric 3D references, and the generated frames provide tracked 2D observations. The relative motion of the sparse anchor set is estimated by OpenCV \texttt{solvePnPRansac} \cite{fischler1981random,lepetit2009epnp}:
\begin{equation}
    \Delta\mathbf{T}_{c,n}^{raw,(t)}
    =
    \argmin_{\Delta\mathbf{T}\in SE(3)}
    \sum_{i\in\mathcal{I}_{n,t}}
    \rho
    \left(
    \left\|
    \mathbf{q}_{i,n}^{(t)}
    -
    \Pi
    \left(
    \mathbf{K},
    \Delta\mathbf{T}
    \bar{\mathbf{p}}_{i}^{c,0}
    \right)
    \right\|_2^2
    \right),
    \label{eq:method_pnp_ransac}
\end{equation}
where $\mathcal{I}_{n,t}$ is the RANSAC inlier set and $\rho(\cdot)$ is the robust loss induced by inlier selection. After solving PnP/RANSAC, the reprojected anchor is
\begin{equation}
    \hat{\mathbf{q}}_{i,n}^{(t)}
    =
    \Pi
    \left(
    \mathbf{K},
    \Delta\mathbf{T}_{c,n}^{raw,(t)}
    \bar{\mathbf{p}}_{i}^{c,0}
    \right).
    \label{eq:projected_anchor}
\end{equation}
Here, for a homogeneous 3D point $[x,y,z,1]^{\top}$, the projection function is defined as
$\Pi(\mathbf{K},[x,y,z,1]^{\top})
=
[f_xx/z+c_x,\; f_yy/z+c_y]^{\top}$.
The RANSAC inlier threshold is set to approximately 8 pixels. The estimated camera-frame motion is transformed to the robot base frame using Eq.~\eqref{eq:relative_motion_base}. The recovered motion is therefore a relative sparse anchor-set motion rather than a CAD-defined canonical object pose.

\paragraph{Reprojection-based candidate selection.}
For each valid frame, rigid-geometric consistency is measured by the mean reprojection residual over RANSAC inliers:
\begin{equation}
\begin{aligned}
    e_{n,t}
    &=
    \frac{1}{|\mathcal{I}_{n,t}|}
    \sum_{i\in\mathcal{I}_{n,t}}
    \left\|
    \mathbf{q}_{i,n}^{(t)}
    -
    \hat{\mathbf{q}}_{i,n}^{(t)}
    \right\|_2,\\
    \bar{e}_{n}
    &=
    \frac{1}{|\mathcal{T}_{n}^{valid}|}
    \sum_{t\in\mathcal{T}_{n}^{valid}} e_{n,t},
    \quad
    e_{n}^{max}
    =
    \max_{t\in\mathcal{T}_{n}^{valid}} e_{n,t}.
\end{aligned}
    \label{eq:method_reprojection_statistics}
\end{equation}
A candidate is accepted only if it satisfies the trajectory-level consistency constraints:
\begin{equation}
\begin{aligned}
    \mathcal{A}
    =
    \big\{
    n \;\big|\;&
    \bar{e}_{n}<\epsilon_{mean},\;
    e_{n}^{max}<\epsilon_{max},\\
    &
    r_n^{valid}>\eta_v,\;
    L_n^{invalid}<L_{max}
    \big\}.
\end{aligned}
    \label{eq:candidate_acceptance_set}
\end{equation}
In our implementation, $\epsilon_{mean}=5$ px and $\epsilon_{max}=12$ px. If no candidate satisfies these constraints, no robot trajectory is executed.

If multiple candidates pass the consistency check, the candidate with the lowest combined score is selected:
\begin{equation}
\begin{aligned}
    C_n
    &=
    \bar{e}_{n}
    +
    \lambda_{max}e_{n}^{max}
    -
    \lambda_v r_n^{valid}
    +
    \lambda_s E_n^{smooth},\\
    n^{*}
    &=
    \arg\min_{n\in\mathcal{A}} C_n ,
\end{aligned}
    \label{eq:candidate_score_selection}
\end{equation}
where $E_n^{smooth}$ penalizes temporally unstable recovered motion. The accepted raw motion sequence is smoothed before execution. Translation is smoothed by a moving-window or Savitzky--Golay filter, while rotation is smoothed in the unit-quaternion representation after sign alignment. Short invalid intervals are filled by $SE(3)$ interpolation only after candidate-level acceptance, and these interpolated frames remain marked as non-verified in diagnostic statistics.

\subsection{Stage 3: Grasp-Conditioned Execution}\label{sec:stage3}

After a generated candidate passes rigid-geometric consistency verification, Stage~3 transfers the verified relative motion to the robot. This stage contains Phase~4, mask-constrained grasp selection and TCP induction, and Phase~5, bounded depth compensation. Its purpose is to connect the verified relative $SE(3)$ motion to a feasible robot TCP trajectory under the selected grasp and local depth feedback.

\subsubsection{Phase 4: Grasp Selection and TCP Induction}\label{sec:grasp_tcp_induction}

AnyGrasp \cite{fang2023anygrasp} generates 6-DoF grasp candidates from the real RGB-D point cloud. Candidate grasps whose contact regions fall outside the task-relevant object mask are removed. Let $\mathcal{J}$ denote the mask-valid candidate set. The final grasp is selected by
\begin{equation}
\begin{aligned}
    \mathcal{H}
    &=
    \left\{
    \left(
    \mathbf{T}_{b,tcp,j}^{(0)},s_j
    \right)
    \right\}_{j=1}^{N_g},\\
    S_j
    &=
    s_j
    -
    \lambda_m E_m(j)
    -
    \lambda_r E_r(j)
    -
    \lambda_c E_c(j),\\
    j^{*}
    &=
    \arg\max_{j\in\mathcal{J}} S_j,
    \quad
    \mathbf{T}_{b,tcp}^{(0)}
    =
    \mathbf{T}_{b,tcp,j^{*}}^{(0)} .
\end{aligned}
    \label{eq:grasp_selection}
\end{equation}
Here, $s_j$ is the grasp confidence score, and $E_m(j)$, $E_r(j)$, and $E_c(j)$ denote mask inconsistency, reachability or near-singularity penalty, and collision penalty, respectively.

Let $\Delta\bar{\mathbf{T}}_{b}^{(t)}$ denote the accepted and smoothed relative motion in the robot base frame. The nominal robot TCP trajectory is induced by applying this motion to the selected real grasp-time TCP pose:
\begin{equation}
    \tilde{\mathbf{T}}_{b,tcp}^{(t)}
    =
    \Delta\bar{\mathbf{T}}_{b}^{(t)}
    \mathbf{T}_{b,tcp}^{(0)},
    \quad
    t=1,\ldots,T.
    \label{eq:method_tcp_transfer}
\end{equation}
This grasp-conditioned formulation preserves the verified relative task motion while anchoring execution to the actual grasp geometry. Therefore, the TCP trajectory is not expected to overlap with the recovered object trajectory or maintain a constant coordinate-wise offset from the object center.

Before execution, the induced trajectory is checked by PyBullet IK with the RM75 URDF. Each frame is initialized from the previous joint solution to improve temporal continuity. A frame is feasible only if the FK back-check satisfies the position threshold, orientation threshold, and all joint limits. In our implementation, the thresholds are 0.01~m and 0.40~rad. The resulting joint trajectory is smoothed by CubicSpline interpolation and Savitzky--Golay filtering, followed by another FK back-check:
\begin{equation}
    \bar{\mathcal{T}}_{tcp}
    =
    \left\{
    \bar{\mathbf{T}}_{b,tcp}^{(t)}
    \right\}_{t=1}^{T}
    =
    \mathcal{S}_{tcp}
    \left(
    \left\{
    \tilde{\mathbf{T}}_{b,tcp}^{(t)}
    \right\}_{t=1}^{T}
    \right),
    \label{eq:ik_validated_tcp_trajectory}
\end{equation}
where $\mathcal{S}_{tcp}(\cdot)$ denotes IK validation, joint-space smoothing, and FK back-checking.

\subsubsection{Phase 5: Bounded Depth Compensation}\label{sec:depth_compensation}

During execution, Phase~5 applies bounded RealSense depth feedback around the IK-validated nominal TCP trajectory:
\vspace{-1mm}
\begin{equation}
\begin{aligned}
    e_d^{(t)}
    &=
    d_{rs}^{(t)}
    -
    d_{exp}^{(t)},\\
    \mathbf{r}_{b}
    &=
    \mathbf{R}_{b,c}
    \begin{bmatrix}
    0 & 0 & 1
    \end{bmatrix}^{\top},\\
    \mathbf{p}_{cmd}^{(t)}
    &=
    \bar{\mathbf{p}}_{tcp}^{(t)}
    +
    \clip
    \left(
    \lambda_d e_d^{(t)},
    -d_{max},
    d_{max}
    \right)
    \mathbf{r}_{b}.
\end{aligned}
    \label{eq:method_depth_correction}
\end{equation}
\vspace{-1mm}
Here, $d_{exp}^{(t)}$ is obtained by projecting the nominal TCP position into the camera frame, and $d_{rs}^{(t)}$ is the median valid depth in the object mask or grasp-neighborhood region. The vector $\mathbf{r}_b$ is the camera depth direction expressed in the robot base frame. In our experiments, $\lambda_d\in[0.3,0.5]$ and $d_{max}\in[0.005,0.01]$~m. The correction modifies only the TCP position along the camera depth direction, while the nominal orientation is preserved. The final command pose is
\begin{equation}
    \mathbf{T}_{cmd}^{(t)}
    =
    \begin{bmatrix}
    \bar{\mathbf{R}}_{tcp}^{(t)}
    &
    \mathbf{p}_{cmd}^{(t)}\\
    \mathbf{0}^{\top}
    &
    1
    \end{bmatrix},
    \label{eq:final_tcp_command_matrix}
\end{equation}
which corresponds to the final trajectory element in Eq.~\eqref{eq:final_command_trajectory}. If no reliable local depth measurement is available, the correction term is set to zero.

This module should be understood as bounded single-axis compensation on top of an offline verified nominal trajectory, rather than as a closed-loop tracking controller or task-level replanner. It can reduce small depth mismatch caused by generated-video geometry, RGB-D noise, calibration residuals, and minor contact disturbance. However, it does not compensate for arbitrary lateral displacement in the camera $X$--$Y$ plane. Small image-plane deviations are tolerated only within the physical margin of the grasp and task. If the object or target region moves substantially during execution, especially in the image plane, the current trajectory is regarded as invalid and the system must reinitialize from a new RGB-D observation and regenerate or re-verify the motion.

\section{Experiments}\label{sec:experiments}

This section evaluates GenVid2Robot on a real RM75 robot through task success, controlled baseline comparisons, generated-candidate filtering, runtime, ablations, trajectory diagnostics, and failure modes.

\begin{figure}[!t]
    \centering
    \includegraphics[width=\columnwidth]{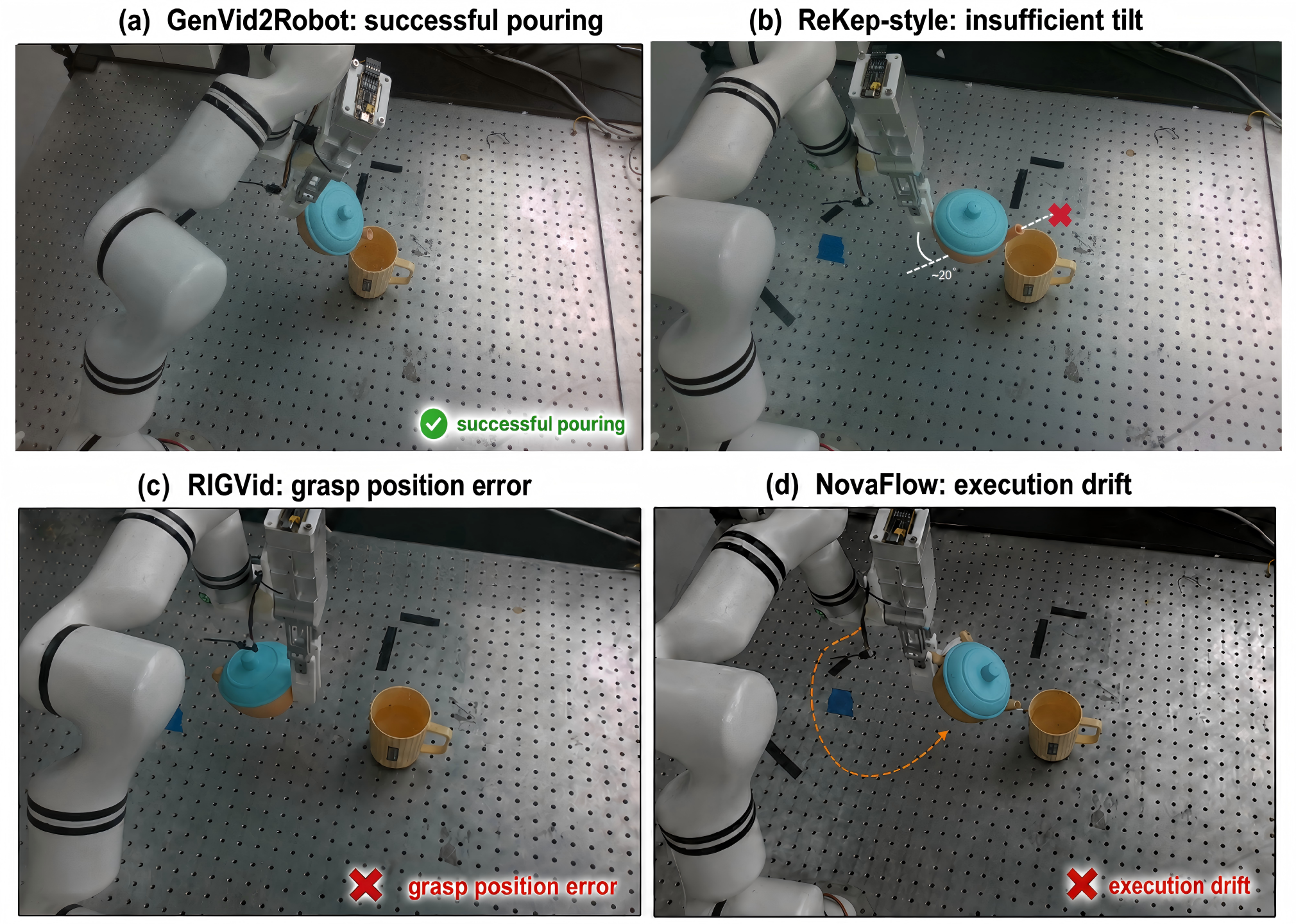}
    \caption{Qualitative pouring comparison. GenVid2Robot completes stable pouring, while the compared variants fail because of insufficient tilt, grasp-alignment error, or trajectory drift.}
    \label{fig:main_comparison}
\end{figure}

\subsection{Experimental Setup and Baselines}\label{sec:experimental_setup}

Experiments are conducted on a real RM75 robot equipped with a parallel gripper and an Intel RealSense RGB-D camera. We evaluate four tabletop manipulation tasks: Pouring, Lifting, Tool Delivery, and Sweeping, with 20 real-robot trials per task. A trial is successful if the robot completes the intended manipulation without losing the object, violating the target interaction, or producing unrecoverable execution drift.

We compare GenVid2Robot with three reproduced baseline variants under the same robot, RGB-D camera, object set, AnyGrasp-based grasp source, IK feasibility checker, and execution interface. ReKep-style execution represents relational keypoint constraint optimization, RIGVid-style execution represents generated-video 6D object-trajectory retargeting, and NovaFlow-style execution represents dense actionable-flow-based retargeting. GenVid2Robot differs by accepting generated motion only when tracked semantic anchors are explainable by first-frame RGB-D anchors under a sparse relative $SE(3)$ model, and by inducing the TCP trajectory from the selected real grasp-time pose.

\subsection{Main Results and Consistency Filtering}\label{sec:main_results}

Fig.~\ref{fig:main_comparison} shows representative pouring behaviors. GenVid2Robot completes stable pouring by combining generated video motion priors, sparse rigid-geometric verification, mask-constrained grasping, and grasp-conditioned TCP induction. Under the same setup, ReKep-style execution tends to under-specify continuous rotation, RIGVid-style execution is sensitive to visual drift and grasp-alignment error, and NovaFlow-style execution may accumulate trajectory drift without explicit sparse rigid $SE(3)$ consistency.

\begin{figure*}[!t]
    \centering
    \includegraphics[width=0.92\textwidth]{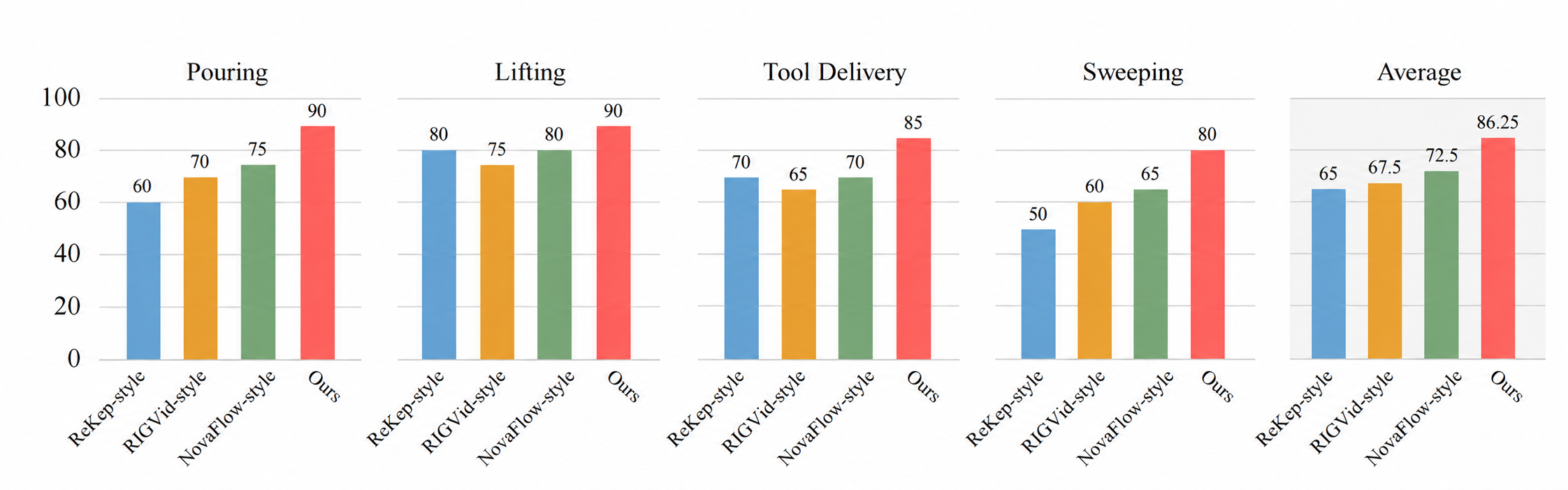}
    \caption{Real-world success rates over 20 trials per task. All variants use the same RM75 robot setup, AnyGrasp-based grasp policy, IK checking, and execution interface. GenVid2Robot achieves the highest success rate across all tasks, indicating the benefit of sparse rigid-geometric consistency and grasp-conditioned TCP induction.}
    \label{fig:success_rate}
\end{figure*}

Fig.~\ref{fig:success_rate} reports the real-robot success rates over 20 trials per task. GenVid2Robot achieves the highest success rate across Pouring, Lifting, Tool Delivery, and Sweeping. Since all variants share the same grasp policy, feasibility checker, and execution interface, the performance difference mainly reflects the motion-transfer mechanism rather than differences in grasp detection or low-level control. The advantage is more evident in Pouring and Sweeping, where continuous rotation, functional alignment, and long-horizon drift strongly affect task completion.

\begin{table}[!t]
    \centering
    \caption{Generated-trajectory filtering.}
    \label{tab:trajectory_filter}
    \renewcommand{\arraystretch}{1.1}
    \setlength{\tabcolsep}{4pt}
    \begin{tabular}{lccc}
        \toprule
        \textbf{Task} & \textbf{Avg. samples} & \textbf{Pass rate} & \textbf{Mean reproj. err.} \\
        \midrule
        Pouring & 2.1 & 61\% & 3.4 px \\
        Lifting & 1.6 & 74\% & 2.8 px \\
        Tool Delivery & 1.9 & 68\% & 3.1 px \\
        Sweeping & 2.4 & 55\% & 4.0 px \\
        \bottomrule
    \end{tabular}
\end{table}

Table~\ref{tab:trajectory_filter} reports generated-candidate filtering. Visually plausible generated videos are not always geometrically transferable: candidates with keypoint drift, non-rigid artifacts, severe occlusion, or unstable sparse rigid motion are rejected before robot execution. This filtering supports the central design of GenVid2Robot: generated videos provide uncertain motion hypotheses, and only motions that remain consistent with first-frame RGB-D geometry under a sparse relative $SE(3)$ model are transferred to the robot.

\subsection{Runtime Profiling}\label{sec:runtime_profiling}

We profile the major pre-execution modules on one representative trial with a 153-frame $640{\times}480$ generated video and 18 semantic anchors sampled from three object parts.

\begin{table}[!t]
    \centering
    \caption{Runtime profiling of major pre-execution modules.}
    \label{tab:runtime_profile_measured}
    \renewcommand{\arraystretch}{1.08}
    \setlength{\tabcolsep}{4pt}
    \footnotesize
    \begin{tabular}{lcc}
        \toprule
        \textbf{Module} & \textbf{Device} & \textbf{Time} \\
        \midrule
        Video gen./prep. & Cloud/GPU & 185.5 s \\
        Anchor generation & Cloud/API+GPU & 224.8 s \\
        CoTracker & GPU & 7.41 s \\
        PnP/RANSAC & CPU & 0.204 s \\
        AnyGrasp & GPU & 2.1 s \\
        IK+smoothing & CPU & 0.5 s \\
        \midrule
        Total latency & Mixed & 420.5 s \\
        \bottomrule
    \end{tabular}
\end{table}

Table~\ref{tab:runtime_profile_measured} shows that the current latency is dominated by cloud-based video generation and VLM grounding, which take 185.5~s and 224.8~s, respectively. In contrast, local geometric and robot-side modules are lightweight: CoTracker takes 7.41~s, PnP/RANSAC takes 0.204~s, AnyGrasp takes 2.1~s, and IK checking with smoothing takes 0.5~s. Therefore, future runtime improvement should mainly focus on faster video generation, local or cached VLM grounding, and parallel candidate evaluation, rather than on the rigid-geometric verification itself.

\subsection{Ablation and Trajectory Diagnostics}\label{sec:ablation}

We further analyze the main components of GenVid2Robot, including rigid-geometric consistency filtering, semantic anchor selection, layout perturbation robustness, bounded depth compensation, and the relation between recovered object motion and executed TCP motion.

\begin{figure}[!t]
    \centering
    \includegraphics[width=\columnwidth]{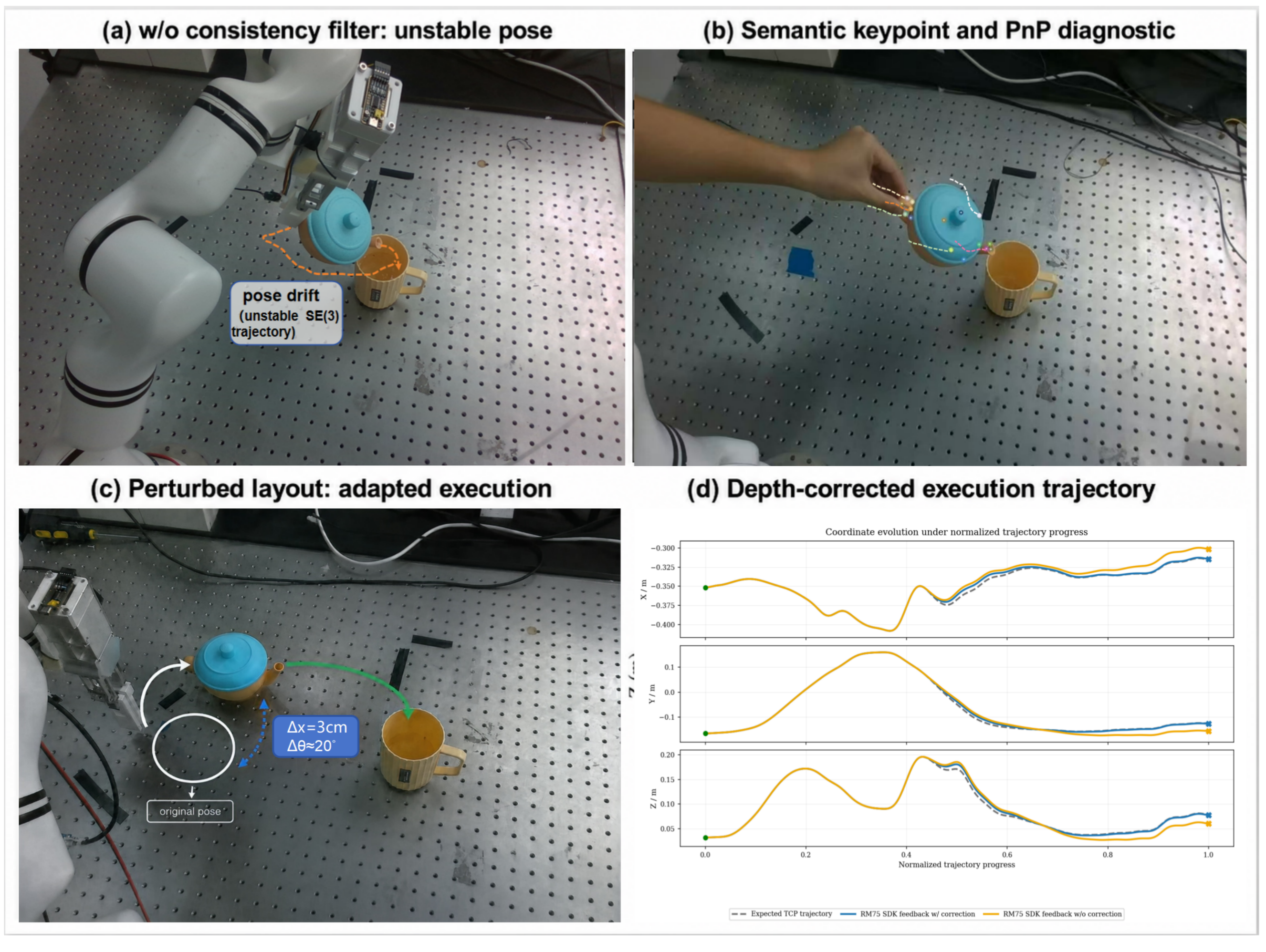}
    \caption{Ablation diagnostics. Rigid-geometric filtering suppresses unstable generated motion, semantic anchors improve PnP stability, layout perturbation is handled by reinitializing first-frame RGB-D anchors, and bounded depth compensation reduces depth-direction execution drift.}
    \label{fig:ablation_diagnostics}
\end{figure}

\begin{table}[!t]
    \centering
    \caption{Ablation of rigid-geometric consistency filtering.}
    \label{tab:filter_ablation}
    \renewcommand{\arraystretch}{1.1}
    \setlength{\tabcolsep}{4pt}
    \begin{tabular}{lcccc}
        \toprule
        \textbf{Setting} & \textbf{Pouring} & \textbf{Lifting} & \textbf{Delivery} & \textbf{Sweeping} \\
        \midrule
        w/o filter & 75\% & 80\% & 70\% & 60\% \\
        GenVid2Robot & \textbf{90\%} & \textbf{90\%} & \textbf{85\%} & \textbf{80\%} \\
        \bottomrule
    \end{tabular}
\end{table}

Table~\ref{tab:filter_ablation} shows that removing rigid-geometric consistency filtering reduces success across all tasks. Without this filter, generated motions with drifting correspondences, non-rigid artifacts, or unstable sparse rigid motion can be transferred to the robot, increasing execution drift and task failure. The effect is more evident in Pouring and Sweeping, where continuous rotation and longer contact-rich motion make the trajectory more sensitive to visual inconsistency.

\begin{table}[!t]
    \centering
    \caption{Anchor selection and PnP diagnostic.}
    \label{tab:keypoint_pnp}
    \renewcommand{\arraystretch}{1.1}
    \setlength{\tabcolsep}{4pt}
    \begin{tabular}{lccc}
        \toprule
        \textbf{Point selection} & \textbf{Survival} & \textbf{Valid PnP} & \textbf{Reproj. err.} \\
        \midrule
        Random points & 72.4\% & 68.1\% & 6.7 px \\
        Uniform mask points & 81.6\% & 78.9\% & 5.1 px \\
        Semantic anchors & \textbf{91.8\%} & \textbf{90.5\%} & \textbf{3.6 px} \\
        \bottomrule
    \end{tabular}
\end{table}

Table~\ref{tab:keypoint_pnp} compares three point-selection strategies using the same tracker, camera intrinsics, PnP/RANSAC solver, and validity rules. Random points are more likely to fall on unstable texture or background regions, while uniform mask points may include geometrically weak or task-irrelevant areas. Part-prompt-guided semantic anchors provide more stable 2D--3D correspondences, resulting in higher survival, higher valid-PnP rate, and lower reprojection error.

\begin{figure*}[!t]
    \centering
    \begin{minipage}[t]{0.50\textwidth}
        \centering
        \includegraphics[width=\linewidth]{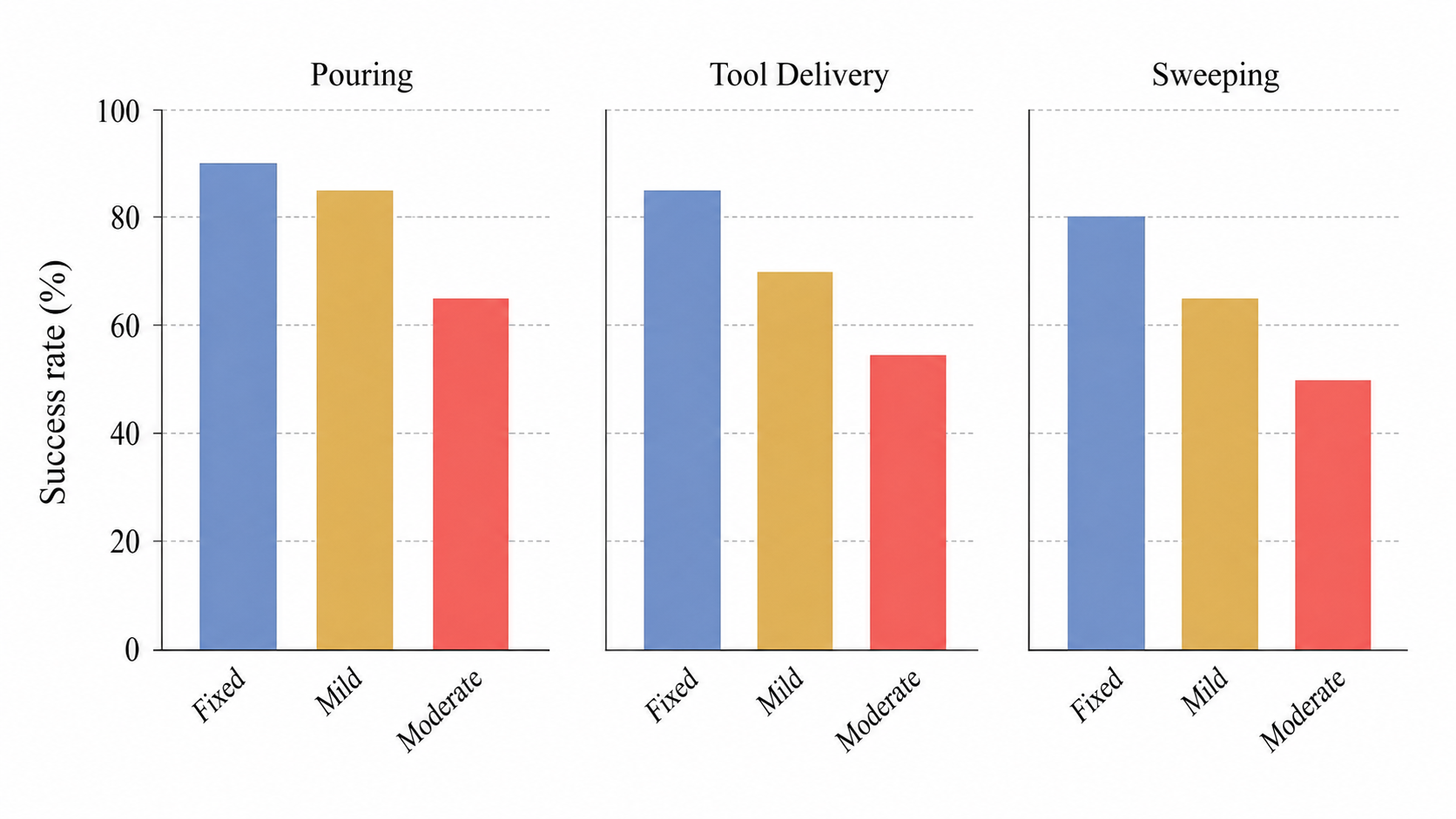}
        \vspace{1mm}
        {\footnotesize (a) Initial layout perturbation robustness.}
    \end{minipage}
    \hfill
    \begin{minipage}[t]{0.4\textwidth}
        \centering
        \includegraphics[width=\linewidth]{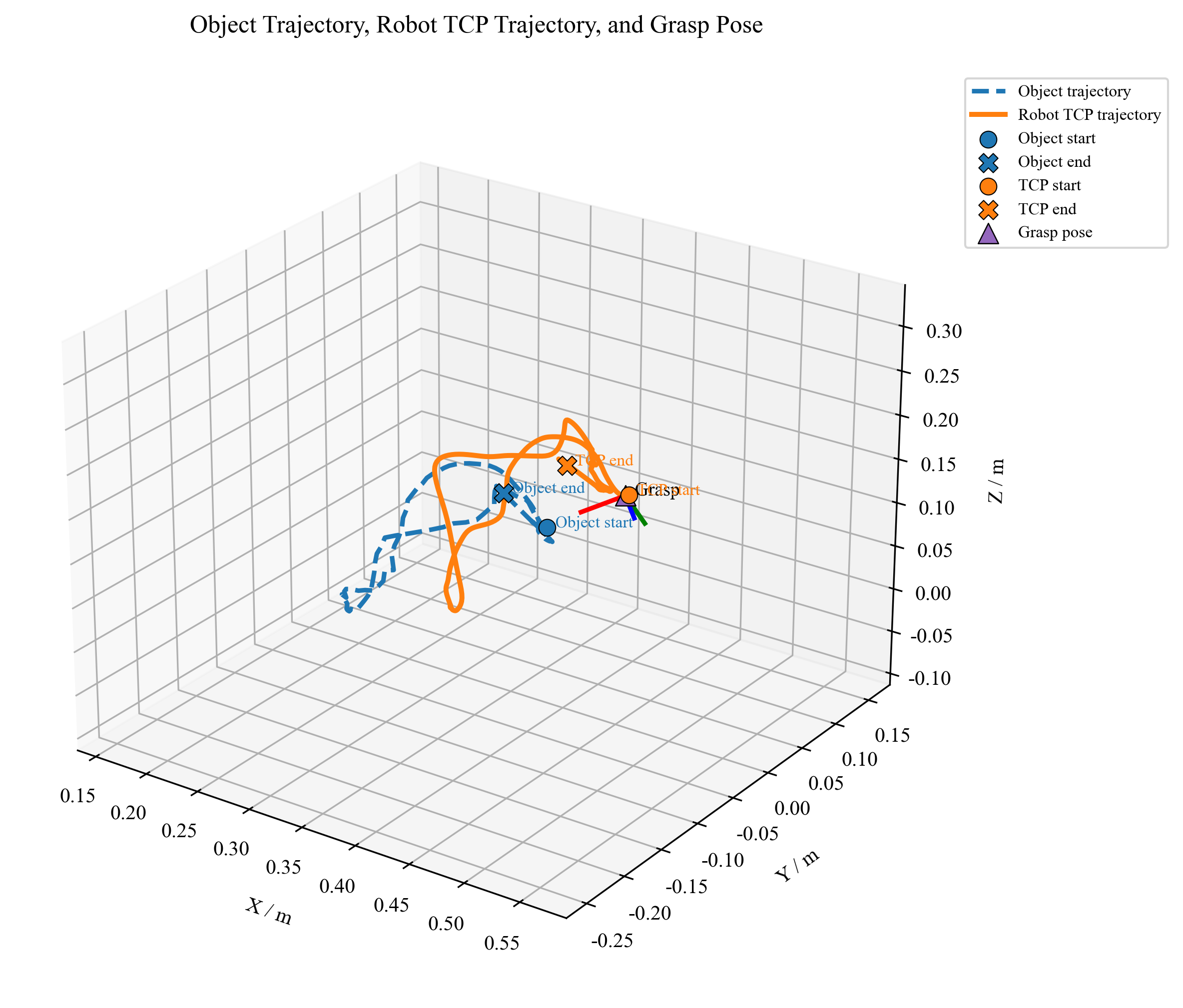}
        \vspace{1mm}
        {\footnotesize (b) Recovered object motion and executed TCP trajectory.}
    \end{minipage}
    \caption{Additional diagnostic results. (a) Success rates under fixed, mild, and moderate initial layout perturbations. (b) Recovered object motion and executed RM75 TCP trajectory in the pouring task. The object and TCP trajectories are not expected to overlap because the TCP is induced from the verified relative motion and the real grasp-time TCP pose.}
    \label{fig:perturb_traj}
\end{figure*}

Fig.~\ref{fig:ablation_diagnostics} and Fig.~\ref{fig:perturb_traj}(a) further show that GenVid2Robot remains robust under moderate layout perturbation because it reconstructs first-frame anchors from the current RGB-D observation and induces the TCP trajectory from the current grasp-time pose, rather than replaying a fixed object-center path. Fig.~\ref{fig:ablation_diagnostics}(d) also supports the bounded depth compensation module: the compensated trajectory remains closer to the expected TCP trajectory than the uncompensated RM75 SDK feedback. This correction is intentionally bounded and single-axis, and should not be interpreted as full online visual servoing.

Fig.~\ref{fig:perturb_traj}(b) visualizes the recovered object motion and executed RM75 TCP trajectory in a pouring trial. This visualization is used for phase-level interpretation rather than coordinate-wise object--TCP tracking error. The recovered object motion provides the relative task motion, while the executed TCP trajectory also reflects the selected grasp geometry, robot kinematics, IK smoothing, and bounded depth correction. Therefore, their consistency should be interpreted in terms of motion phase, direction, and task intent, rather than point-wise spatial coincidence.

\subsection{Failure Analysis}\label{sec:failure_analysis}

Across 80 GenVid2Robot trials, 69 trials succeed and 11 trials fail. The failures consist of eight workspace-limit failures, two excessive keypoint-loss failures, and one grasp-execution error. The dominant failure type is workspace-limit failure. These cases occur when the generated motion passes rigid-geometric consistency verification, but the induced grasp-conditioned TCP trajectory approaches the reachability boundary, a near-singular posture, or a collision-prone region of the RM75 workspace. This indicates that sparse rigid-geometric consistency can validate the generated visual motion, but it does not by itself guarantee global reachability, manipulability, or collision-free execution along the entire robot trajectory.

The two excessive keypoint-loss failures occur when occlusion, severe appearance change, or ambiguous object boundaries leave too few valid 2D--3D correspondences for stable PnP/RANSAC. In these cases, GenVid2Robot rejects or invalidates the generated motion rather than sending an unstable trajectory to the robot. The remaining grasp-execution error is caused by local depth noise or small contact disturbance during physical interaction. Although mask-constrained grasping and bounded depth compensation improve robustness, they cannot eliminate all contact-level uncertainty. These failure cases suggest that future improvements should combine the proposed rigid-geometric verification with workspace-aware trajectory optimization, manipulability-aware planning, and stronger online grasp or contact feedback.

\section{Discussion}\label{sec:discussion}

The experiments show that generated videos can provide useful visual motion priors, but they must be geometrically and physically grounded before robot execution. ReKep-style relational constraints may under-specify continuous object rotation, RIGVid-style generated-video trajectory recovery is sensitive to visual drift and grasp grounding, and NovaFlow-style actionable flow may accumulate trajectory drift without explicit sparse rigid consistency. In contrast, GenVid2Robot verifies whether generated 2D anchor motion is explainable by first-frame RGB-D anchors under a sparse relative $SE(3)$ model, and transfers only verified motion to the robot through the real grasp-time TCP pose.

A central design choice is to use first-frame sparse RGB-D anchors rather than a CAD-defined object frame. Therefore, the recovered motion should be interpreted as a relative anchor-set motion prior, not as a canonical object 6D pose. The TCP trajectory is induced by applying this relative motion to the selected grasp-time TCP pose, so object and TCP trajectories are not expected to overlap or maintain a constant coordinate-wise offset. This is important for manipulation tasks such as pouring, where the functional motion is object-centric but the executable wrist motion depends on the grasp location and robot kinematics.

The current system still has limitations. It depends on generated-video quality and may reject or fail on sequences with severe deformation, incorrect task semantics, strong occlusion, or insufficient visible anchors. Sparse PnP can also be unstable for textureless, symmetric, or poorly distributed anchor regions. In addition, bounded depth compensation is only a local single-axis correction on top of an offline verified nominal trajectory; it is not full online visual servoing, force-aware control, or task-level replanning. Large target displacement, especially in the camera image plane, remains outside the current capability and requires scene reinitialization. Future work will incorporate stronger video selection, workspace-aware and manipulability-aware trajectory optimization, online grasp refinement, and tactile or force feedback for more robust contact-rich manipulation.

\section{Conclusion}\label{sec:conclusion}

This paper presented GenVid2Robot, a rigid-geometric consistency framework for converting generated video motion priors into executable robot manipulation trajectories. Instead of treating generated videos as robot demonstrations, GenVid2Robot interprets them as uncertain 2D visual motion hypotheses, initializes task-relevant semantic anchors from the real first RGB-D observation, verifies generated anchor tracks through sparse relative $SE(3)$ consistency, and transfers only accepted motion to the robot through mask-constrained grasping and grasp-conditioned TCP induction. Bounded RealSense depth compensation further reduces small execution-time depth mismatch.

Real-robot experiments on the RM75 platform show that GenVid2Robot improves success rates over reproduced ReKep-style, RIGVid-style, and NovaFlow-style variants under controlled settings. The results suggest that generated videos can support robot manipulation when their visual motion is grounded by first-frame metric geometry, grasp feasibility, and robot execution constraints. Current limitations include dependence on generated-video quality, sparse-anchor visibility, workspace reachability, and the lack of full online visual servoing or force-aware replanning.


\newcommand{\bioimg}[1]{%
    \includegraphics[width=1.0in,height=1.0in,clip,keepaspectratio]{#1}%
}

\begin{IEEEbiography}[{\bioimg{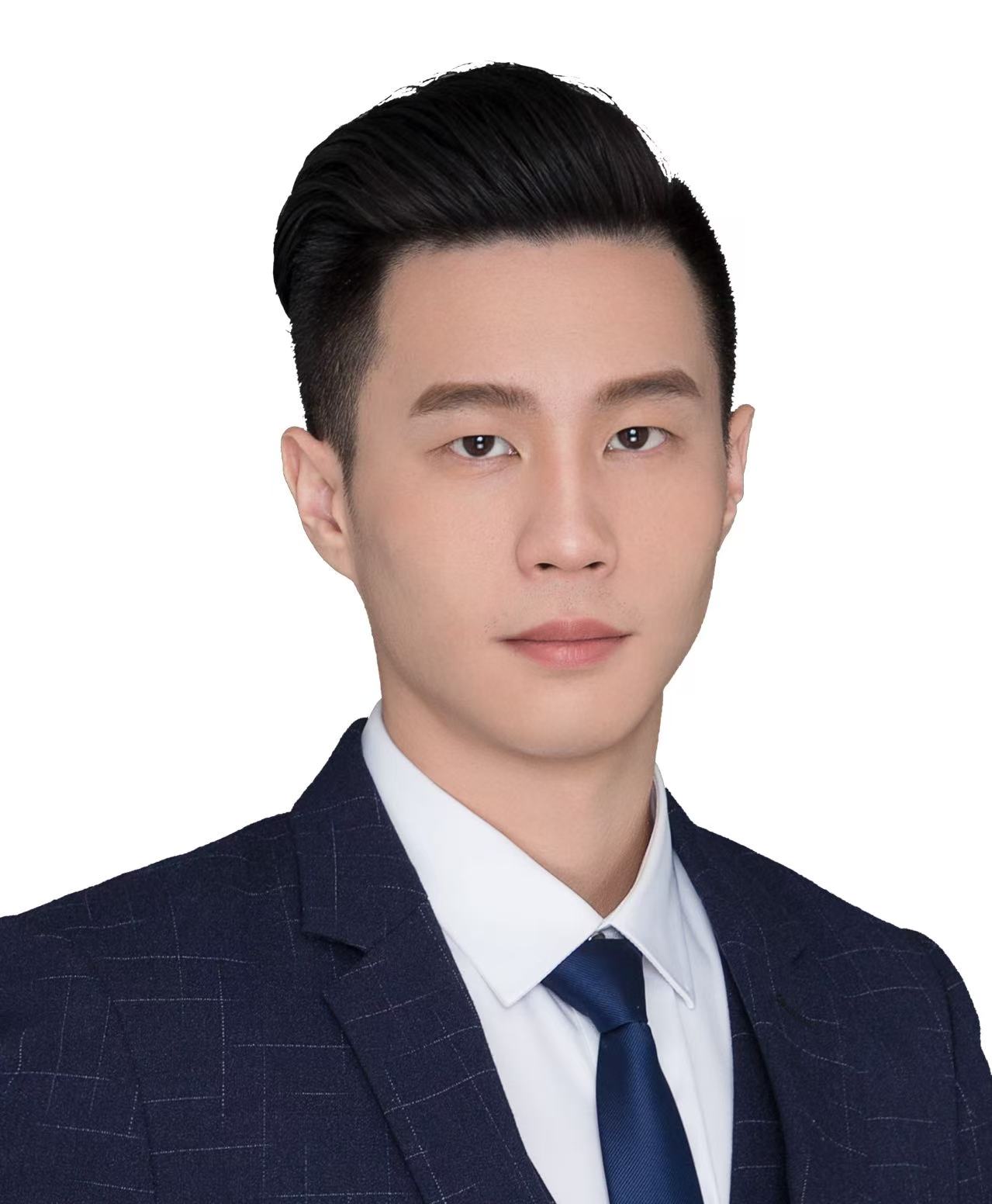}}]{Haohui Huang}
(Member, IEEE) received the B.S. and M.S. degrees from Guangdong University of Technology, Guangzhou, China, in 2011 and 2014, respectively, and the Ph.D. degree in control science and engineering from South China University of Technology, Guangzhou, China, in 2020. He was a postdoctoral researcher in human--robot interaction at Shanghai Jiao Tong University, Shanghai, China. Since 2023, he has been with the School of Automation, Guangdong University of Technology, Guangzhou, China. His research interests include robotics, compliant control, and human--robot interaction.
\end{IEEEbiography}
\vspace{-3.0em}

\begin{IEEEbiography}[{\bioimg{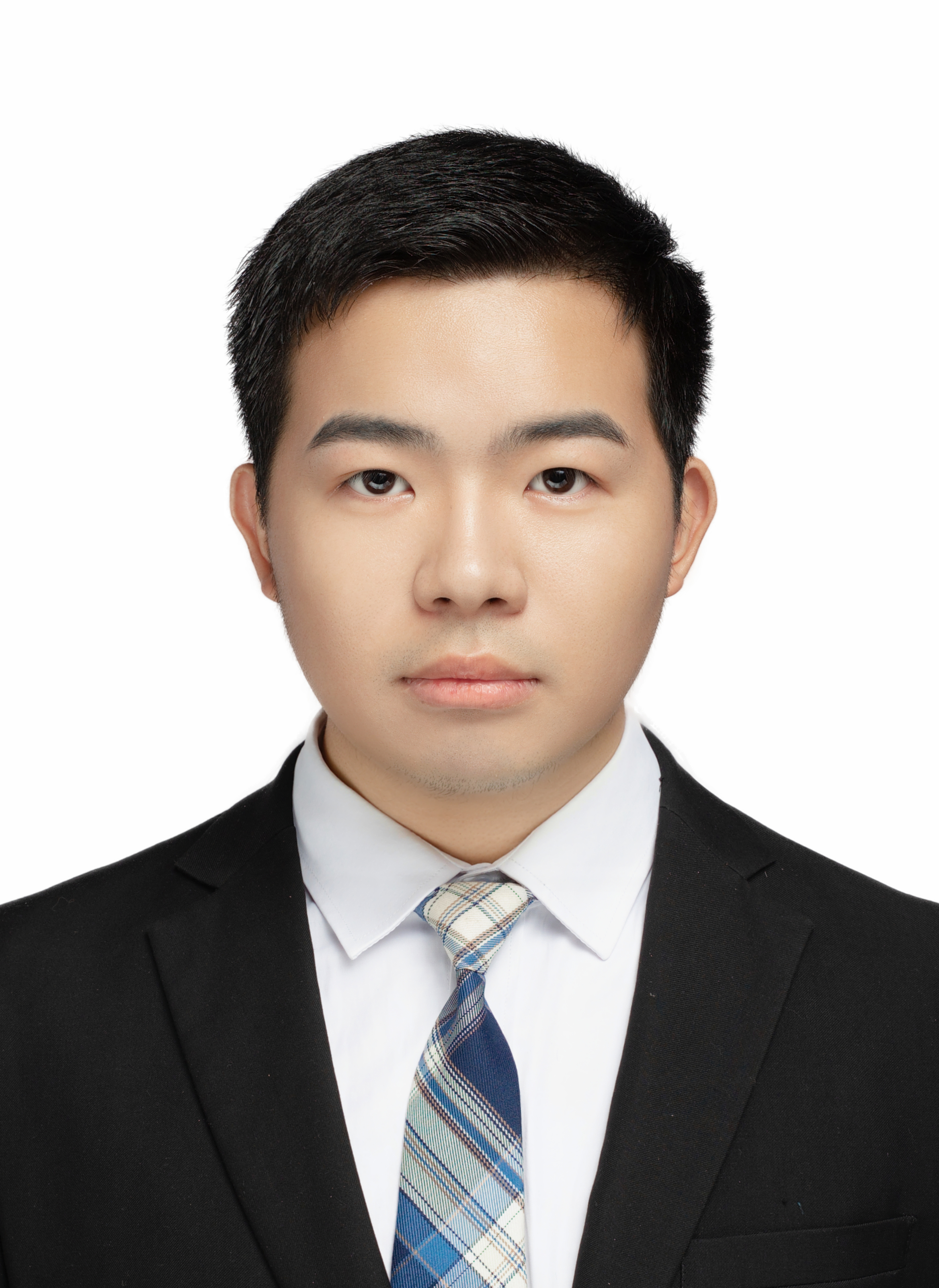}}]{Xi Yuan}
received the B.S. degree from Yangtze University, Jingzhou, China, in 2024. He is currently pursuing the M.S. degree with the School of Automation, Guangdong University of Technology, Guangzhou, China. His research interests include robot manipulation, video-guided robot learning, multimodal perception, and visual motion understanding.
\end{IEEEbiography}
\vspace{-3.0em}

\begin{IEEEbiography}[{\bioimg{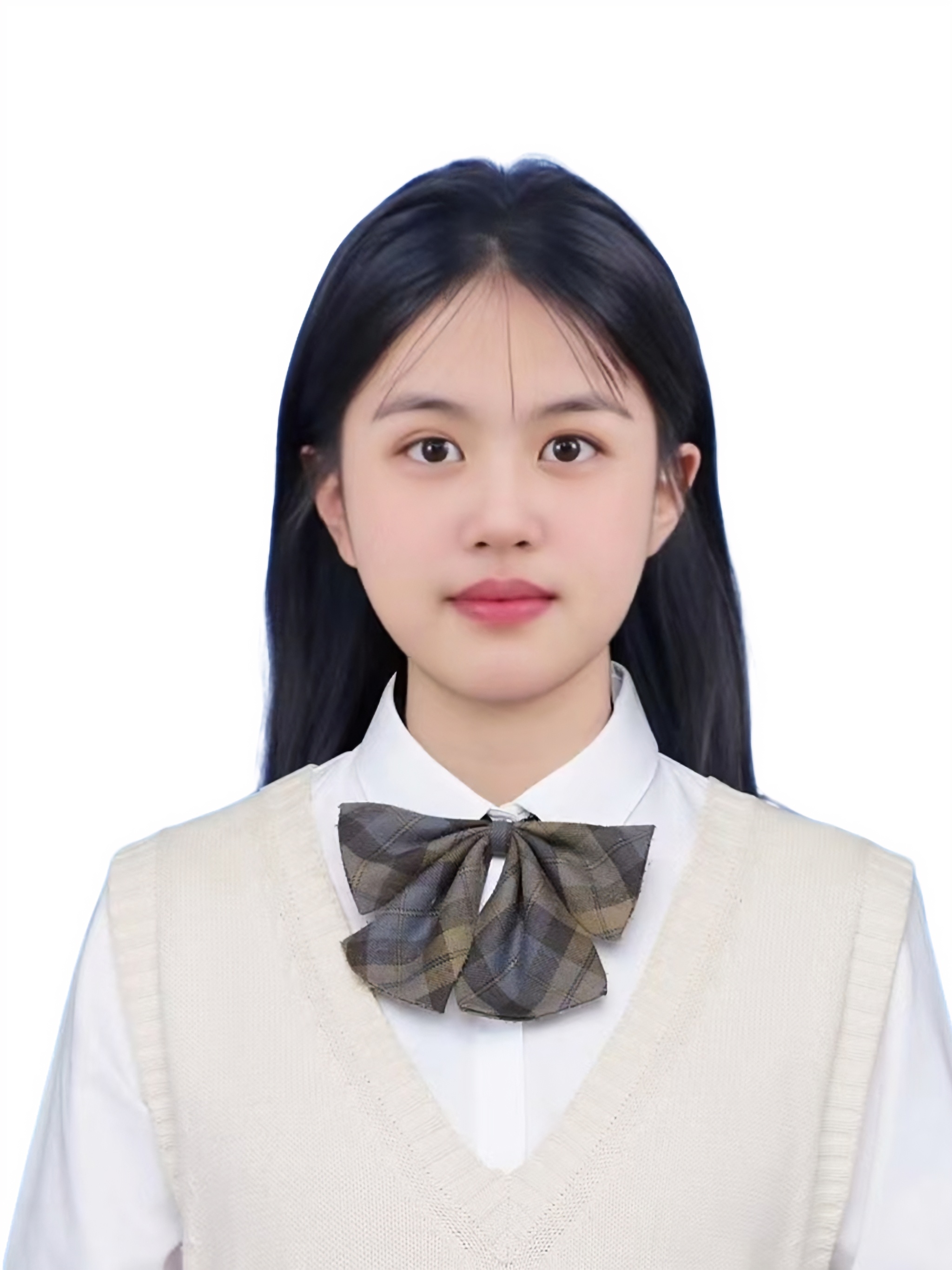}}]{Panpan Liao}
is currently an undergraduate student with the School of Automation, Guangdong University of Technology, Guangzhou, China. The research interests include robot perception, robot manipulation, and vision-based robot learning.
\end{IEEEbiography}
\vspace{-3.0em}

\begin{IEEEbiography}[{\bioimg{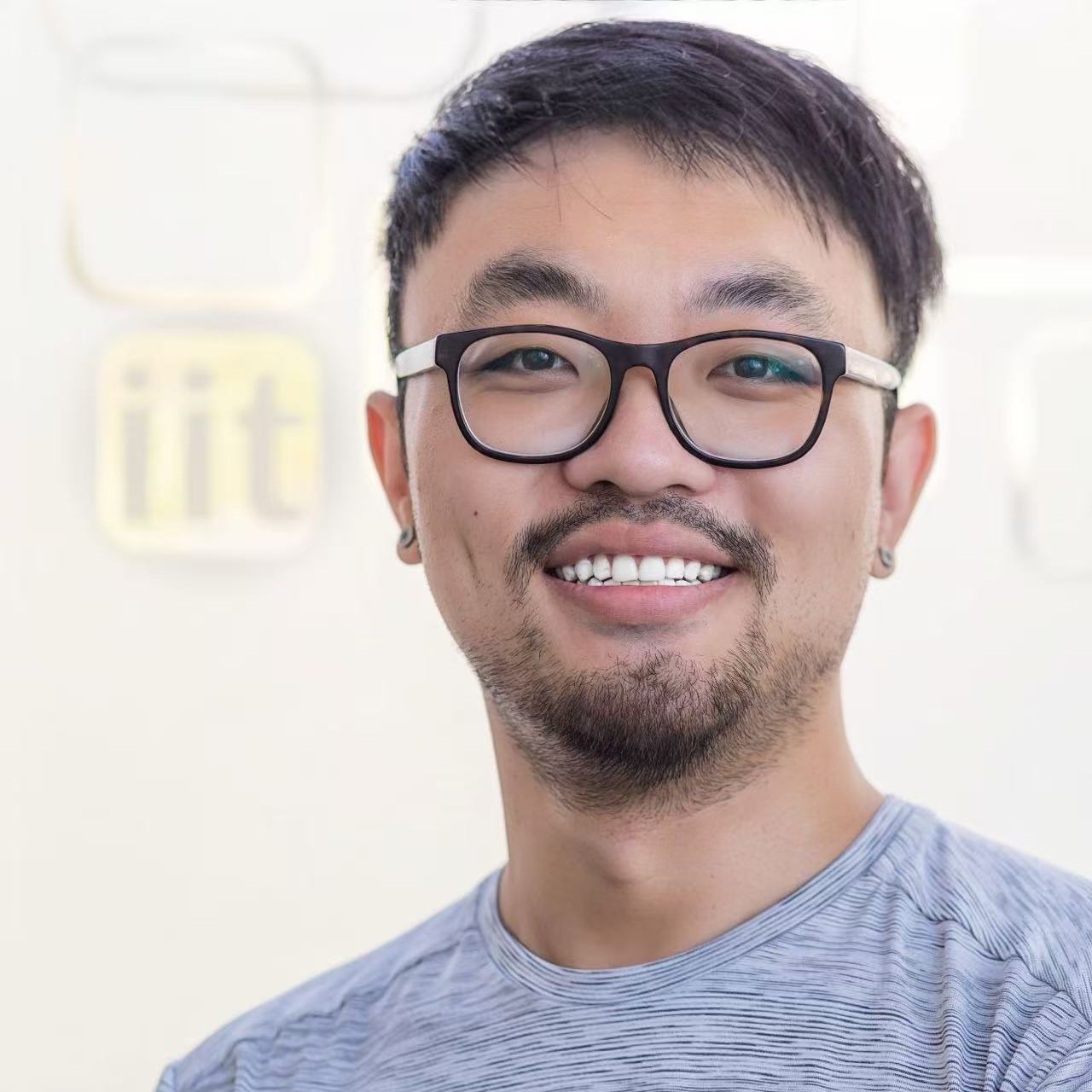}}]{Tao Teng}
(Member, IEEE) received the B.S. degree in automation and the M.S. degree in pattern recognition and intelligent systems from South China University of Technology, Guangzhou, China, in 2016 and 2019, respectively, and the Ph.D. degree in agri-food systems (robotics) from Universit\`a Cattolica del Sacro Cuore, Italy, in affiliation with the Istituto Italiano di Tecnologia, Italy, in 2023. He was a postdoctoral researcher in robotics with the Hong Kong Centre for Logistics Robotics and The Chinese University of Hong Kong, Hong Kong, China. He is currently with the University of Liverpool, Liverpool, U.K. His research interests include human--robot interaction, robot learning, and control.
\end{IEEEbiography}
\vspace{-3.0em}

\begin{IEEEbiography}[{\bioimg{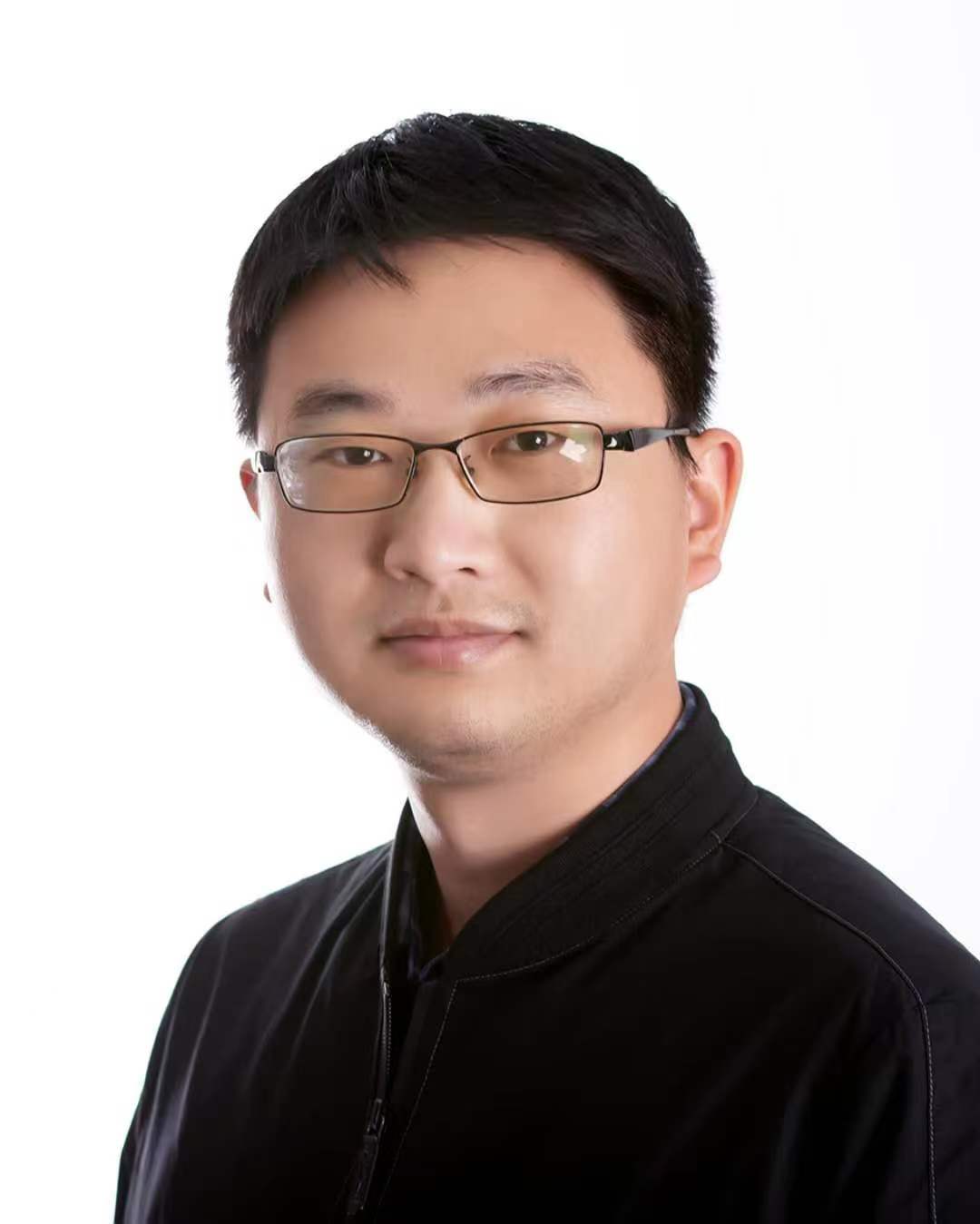}}]{Chenguang Yang}
(Fellow, IEEE) received the B.Eng. degree from Northwestern Polytechnical University, Xi'an, China, in 2005, and the Ph.D. degree from the National University of Singapore, Singapore, in 2010. He is currently a Professor with the Department of Computing, The Hong Kong Polytechnic University. He previously held academic positions at the University of Liverpool, the University of the West of England, and South China University of Technology. He received the IEEE Transactions on Robotics Best Paper Award in 2012 and the IEEE Transactions on Neural Networks and Learning Systems Outstanding Paper Award in 2022. His research interests include robot control and learning, human--robot interaction, and intelligent system design.
\end{IEEEbiography}
\vspace{-3.0em}

\begin{IEEEbiography}[{\bioimg{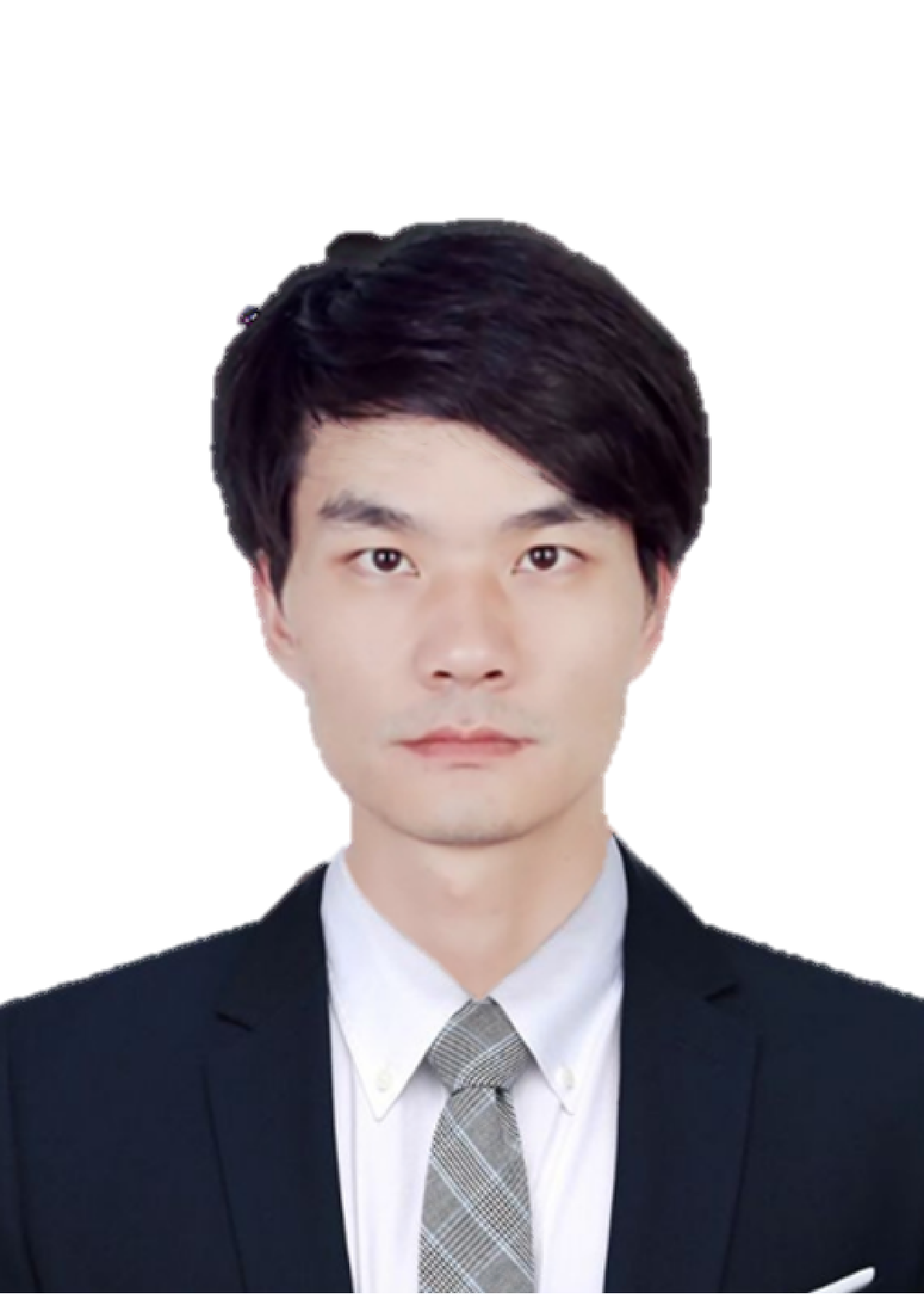}}]{Jing Guo}
(Member, IEEE) received the B.S. and M.S. degrees from Guangdong University of Technology, Guangzhou, China, in 2009 and 2012, respectively, and the Ph.D. degree from LIRMM, CNRS--University of Montpellier, Montpellier, France, in 2016. He was a Research Fellow with the National University of Singapore, Singapore, from 2016 to 2018. He is currently an Associate Professor with Guangdong University of Technology, Guangzhou, China. His current research interests include robotic control and learning, haptic bilateral teleoperation, and surgical robotics. He has served as a Guest Editor for \emph{IEEE Robotics and Automation Letters} and \emph{Frontiers in Robotics and AI}.
\end{IEEEbiography}
\vspace{-3.0em}

\begin{IEEEbiography}[{\bioimg{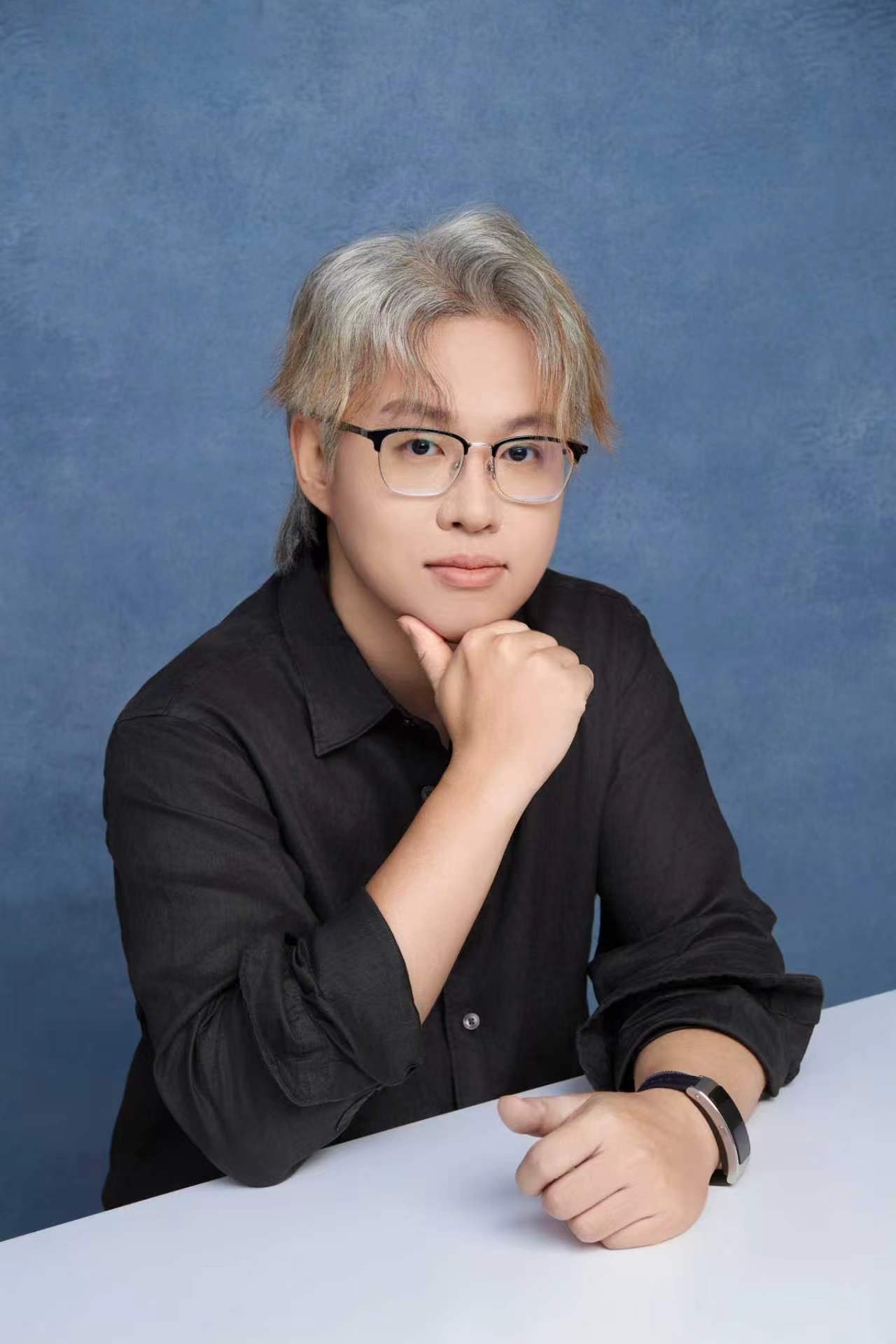}}]{Yi Guo}
(Member, IEEE) received the B.S. degree in electrical and computer engineering from Shanghai Jiao Tong University, Shanghai, China, and the Ph.D. degree in computer science and engineering from the Hong Kong University of Science and Technology, Hong Kong, China. He is currently an Associate Researcher with the State Key Laboratory of Submarine Geoscience, School of Automation and Intelligent Sensing, Shanghai Jiao Tong University, Shanghai, China. His research interests include embodied AI, multimodal perception, and decision-making for robotic systems. He is also a member of ACM.
\end{IEEEbiography}

\end{document}